\def\year{2020}\relax
\documentclass[letterpaper]{article} 
\usepackage{aaai20}  
\usepackage{times}  
\usepackage{helvet} 
\usepackage{courier}  
\usepackage[hyphens]{url}  
\usepackage{graphicx} 
\usepackage{graphicx}  
\usepackage{amsmath} \usepackage{amssymb}
\usepackage{fancybox}
\usepackage{tikz}
\usetikzlibrary{arrows,shapes,calc}
\usepackage[tworuled,norelsize]{algorithm2e}
\usepackage{empheq}
\usepackage{paralist}
\usepackage{pgfplots}
\usepackage{float}
\usepackage{adjustbox}
\usepackage{multirow}
\usepackage{subfig}

\newcommand{\PCA}{\text{\sf PCA}}
\newcommand{\RPCA}{\text{\sf RPCA}}

\newcommand{\refe}[1]{(\ref{#1})}

\makeatletter
\newcommand{\removelatexerror}{\let\@latex@error\@gobble}
\makeatother

\usepackage{xcolor}

\title{A Bias Trick for Centered Robust Principal Component Analysis}
\author{Baokun He, Guihong Wan, and Haim Schweitzer\\
\{Baokun.He, GuiHong.Wan, Haim\}@utdallas.edu\\
The University of Texas at Dallas\\
800 W. Campbell Road, Richardson, Texas
}

\begin{document}
\maketitle
\begin{abstract}
Outlier based Robust Principal Component Analysis (RPCA) requires centering of 
the non-outliers.
We show a ``bias trick'' that automatically centers these non-outliers.
Using this bias trick we obtain the first  RPCA algorithm that is optimal with 
respect to centering.
\end{abstract}

\section{Introduction}
Principal Component Analysis (PCA)
is arguably the most widely used dimension reduction technique.
It is known that the PCA model is heavily influenced by data outliers.
The detection and removal of such outliers is a key component of
robust variants of the PCA.

There are two main variants of standard PCA:
centered and uncentered.
The only difference between them is that in centered PCA
there is a preliminary step where the data is being centered.
From a computational point of view, there is little difference between
these variants.
For this reason, most recently published fast algorithms
for computing PCA ignore the centering of the data.
The situation is very different for algorithms that attempt to compute
RPCA by identifying some points as outliers to be removed.
The problem is that the centering should be applied only
to the non-outliers, but they are unknown.

Some previously proposed RPCA algorithms
perform initial centering of the data
but do not update the center based on the outliers.
These include~\cite{Zhang15,Xu10,astarRPCA}.
Other algorithms such as~\cite{OutierRobust15,Rahmani17}
do not explicitly center the data.
The first assumes a probability distribution of the mean,
and the second considers only directions of vectors which makes
centering unnecessary.
Other algorithms such as~\cite{robpca} handle the centering as part
of the algorithm,
but not optimally.
This review of the current state of the art suggests that
optimal centering in RPCA is not fully solved.

We propose a general method (a bias trick) that can be used to convert
{\em any}
robust algorithm that does not perform centering into an algorithm
that performs centering optimally.
In fact, the bias trick can be used to convert any algorithm that computes
uncentered PCA into an algorithm that computes a centered PCA.

Using the bias trick with the
algorithm of ~\cite{astarRPCA} that computes optimal uncentered RPCA
gives the first optimal centered RPCA algorithm.
We implemented this algorithm and describe some experimental results,
showing improved performance over all competitors.

\section{The Bias Trick} \label{sec:trick}
Let $\PCA()$ be an uncentered PCA algorithm.
It gets as input the matrix $X$  of size $m {\times} n$ 
and the number $k$ of desired principal vectors.
It returns the principal vectors as the matrix $V$ of size $m {\times} k$,
and $k$ eigenvalues.
To apply the bias trick and obtain the centered PCA
we do the following:
\begin{compactdesc}
\item[1.] Select a large value $b$. (See Section~\ref{sec:proof}.)
\item[2.] Add $b$ as an additional coordinate to each column of $X$,
  creating a new matrix $X_b$ of size $(m{+}1) {\times} n$.
\item[3.] Run $\PCA()$ on $X_b$
to compute $k{+}1$ eigenvectors and eigenvalues.
Each eigenvector is of size $(m{+}1)$.
\item[4.]
Let $\lambda^b_1 \ldots, \lambda^b_{k+1}$ be the eigenvalues computed
in Step~3. Then the $k$ eigenvalues of the centered PCA are 
approximately
$\lambda^b_2 \ldots \lambda^b_{k{+}1}$.
\item[5.]
Let $u^b_1 \ldots, u^b_{k+1}$ be the eigenvectors computed
in Step~3. 
Let $v_j$ be the the $j$th eigenvector of the centered PCA.
It is given approximately by the top $m$ values of the $u^b_{k{+}1}$.
\end{compactdesc}

\smallskip
\noindent
Clearly, the bias trick is not an improvement over the standard 
centered PCA algorithm. 
It is more costly and less accurate.
But, it has the advantage that it also works for centered RPCA where it does not require
advanced knowledge of the outliers.
Applying the bias trick for computing centered RPCA
can be achieved by using $\RPCA()$ instead of $\PCA()$, where $\RPCA()$ is
any uncentered RPCA algorithm.

\section{Correctness of the Bias Trick}\label{sec:proof}
The following theorem is proved as a corollary in the Appendix.

\begin{paragraph}{Theorem:}
Let $X$ be the data matrix and let $\mu$ be the mean of $X$ columns.
For any desired accuracy of computing the centered PCA of $X$
there exists $0 {<} \epsilon {<} 1$
such that 
setting $b \geq \frac{\sqrt{1-\epsilon^2}}{\epsilon} \| \mu \|$
in the procedure outlined in Section~\ref{sec:trick} gives the
desired approximation.
\end{paragraph}

\section{Experiments}
\begin{figure}
    \centering
    \begin{adjustbox}{width=.65\columnwidth}
\begin{tikzpicture}

\definecolor{color0}{rgb}{0.12156862745098,0.466666666666667,0.705882352941177}

\begin{axis}[
tick align=outside,
tick pos=left,
title={iris},
x grid style={white!69.01960784313725!black},
xlabel={\(\displaystyle \epsilon\)},
xmin=-0.039, xmax=1.039,
xtick style={color=black},
xtick={-0.2,0,0.2,0.4,0.6,0.8,1,1.2},
xticklabels={−0.2,0.0,0.2,0.4,0.6,0.8,1.0,1.2},
y grid style={white!69.01960784313725!black},
ylabel={\(\displaystyle Err_{\text{eigenvalues}}\)},
ymin=-0.0117375202538132, ymax=0.246487938686187,
ytick style={color=black},
ytick={-0.05,0,0.05,0.1,0.15,0.2,0.25},
yticklabels={−0.5,0.0,0.5,1.0,1.5,2.0,2.5}
]
\addplot [semithick, color0]
table {%
0.01 6.07095896244832e-10
0.02 9.71669803820013e-09
0.03 4.9217487039766e-08
0.04 1.55669829879584e-07
0.05 3.80424991105711e-07
0.06 7.89792785761299e-07
0.07 1.46525839036958e-06
0.08 2.50374930818805e-06
0.09 4.01795367196086e-06
0.1 6.13669130793173e-06
0.11 9.00533925068293e-06
0.12 1.27863136223751e-05
0.13 1.76596100783874e-05
0.14 2.38234052881358e-05
0.15 3.14947222248402e-05
0.16 4.09101623453348e-05
0.17 5.23267080779729e-05
0.18 6.60225993884861e-05
0.19 8.2298288574296e-05
0.2 0.000101477477863199
0.21 0.000123908244783591
0.22 0.000149964260813578
0.23 0.000180046109260847
0.24 0.000214582708908634
0.25 0.000254032850565369
0.26 0.000298886854267935
0.27 0.000349668355618447
0.28 0.000406936230486116
0.29 0.000471286668131791
0.3 0.000543355403700167
0.31 0.000623820122040481
0.32 0.000713403045848684
0.33 0.000812873722334808
0.34 0.000923052023859423
0.35 0.00104481137937812
0.36 0.00117908225516286
0.37 0.00132685590473812
0.38 0.00148918841007846
0.39 0.00166720503786243
0.4 0.00186210493698793
0.41 0.00207516620588148
0.42 0.00230775136085614
0.43 0.00256131323958162
0.44 0.00283740137723482
0.45 0.00313766889604747
0.46 0.00346387995330646
0.47 0.0038179177967808
0.48 0.00420179348160755
0.49 0.00461765530769111
0.5 0.00506779904246735
0.51 0.00555467900041413
0.52 0.00608092005740715
0.53 0.00664933068598865
0.54 0.00726291710611204
0.55 0.00792489865514886
0.56 0.0086387244916793
0.57 0.00940809175856218
0.58 0.0102369653438507
0.59 0.0111295993916667
0.6 0.0120905607305428
0.61 0.0131247544035212
0.62 0.0142374515026932
0.63 0.0154343195309546
0.64 0.0167214555360174
0.65 0.0181054222852752
0.66 0.0195932877766436
0.67 0.0211926684083718
0.68 0.0229117761610151
0.69 0.0247594701771369
0.7 0.0267453131581439
0.71 0.0288796330332098
0.72 0.0311735903913067
0.73 0.0336392522036682
0.74 0.0362896723985164
0.75 0.039138979880976
0.76 0.0422024746163025
0.77 0.0454967324082108
0.78 0.0490397190028844
0.79 0.0528509141216251
0.8 0.0569514459615097
0.81 0.0613642365834702
0.82 0.0661141584041875
0.83 0.0712282016771318
0.84 0.0767356523157588
0.85 0.0826682785482556
0.86 0.0890605234517151
0.87 0.0959496978870964
0.88 0.103376163614112
0.89 0.11138348665071
0.9 0.120018518975941
0.91 0.129331311798677
0.92 0.139374615869011
0.93 0.150202341114982
0.94 0.161865890516371
0.95 0.17441135751513
0.96 0.187894070558945
0.97 0.202394062693953
0.98 0.217997884176576
0.99 0.234750417825278
};
\end{axis}

\end{tikzpicture}
    \end{adjustbox}
    \caption{
        Eigenvalue error 
        with the bias trick on ``iris'' as a function of $\epsilon$.
        The error is almost 0 for $\epsilon$ below 0.2.
        }
    \label{fig:wvalues}
\end{figure}
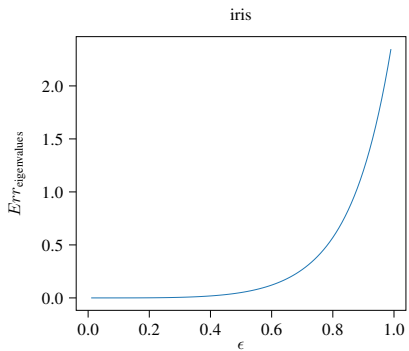

In the first experiment, we demonstrate that
centered PCA implemented with the bias trick
returns accurate eigenvalues and eigenvectors.
The error for the iris data
(from UC Irvine) with various $\epsilon$ values is shown
in Figure~\ref{fig:wvalues}.
Observe that for moderate values of $\epsilon$ and $b$ the error
is almost 0.
Similar results were obtained with other datasets,
suggesting that $\epsilon {\approx} 0.2$, 
or $b {\approx} 5 \| \mu \|$,
may give sufficient accuracy.

\subsubsection{Optimal Centered Robust PCA.}
We describe the results of using the algorithm of~\cite{astarRPCA}
with the bias trick.
The original algorithm computes optimal uncentered RPCA.
With the bias trick the algorithm computes optimal centered RPCA.
To the best of our knowledge this is the first centered algorithm with
guaranteed optimality. We refer to this algorithm as COPT.

Tables~{\ref{tab:outlier_pursuit},\ref{tab:CoP}
show errors for several algorithms
using code provided by the authors.
The values are the average reconstruction error
of the $n$ non-outlier points: 
$E_\text{rpca} {=} \frac{1}{n}
\sum_i{||(x_i {-} \mu)
{-} V_\text{rpca} V_\text{rpca}^T (x_i {-} \mu)||^2}$.
Our COPT is clearly superior.

\begin{table}
\caption{$E_{\text{rpca}}$ comparison to Outlier-Pursuit algorithm }
\label{tab:outlier_pursuit}
\centering
\begin{adjustbox}{width=.9\columnwidth}
\begin{tabular}{|r|r|r|r|r|r|}
\hline
dataset & $k:r$ & bias & COPT & Outlier-Pursuit \\ \hline
smoking & 2:1   & 1753 & 703.55    & 1159.7           \\ \hline
wdbc    & 20:2  & 6632 & 241.460   & 304.24           \\ \hline
wine    & 5:2   & 2244 & 14.7220   & 15.5000          \\ \hline
\end{tabular}
\end{adjustbox}
\end{table}

\begin{table}
\caption{$E_{\text{rpca}}$ comparison to CoP algorithm }
\label{tab:CoP}
\centering
\begin{adjustbox}{width=.9\columnwidth}
\begin{tabular}{|r|r|r|r|r|r|}
\hline
dataset & $k:r$ & bias & COPT & CoP \\ \hline
smoking & 1:1   & 1753 & 1343.0      & 1343.0  \\ \hline
wdbc    & 17:2  & 6632 & 252.14      & 498.75  \\ \hline
wine    & 5:2   & 2244 & 14.7220     & 15.5000 \\ \hline
\end{tabular}
\end{adjustbox}
\end{table}

Figures~\ref{fig:A*outliers}, \ref{fig:outlier_pursuit} compare 
the results of our COPT algorithm
to the results of the Outlier-Pursuit algorithm. 
Five outliers were selected based on the first two principal vectors.
The left panel in both figures shows the location of points in the plane 
defined by these vectors.
The right panel is the location of points on the plane defined by the first and 
third principal vectors.  
The horizontal line is the plane composed of the first two principal vectors.
Indeed, the locations of outliers are far away from the locations of non-outliers
 in the third principal vector direction. 
As shown on the right panel in Figure~\ref{fig:A*outliers}
 the five selected outliers are the furthest ones away from the horizontal line.
 In Figure \ref{fig:outlier_pursuit} the outliers are not the ones farthest away
 especially for the points 69 and 95.
 This illustrates that our COPT returns better outliers than Outlier-Pursuit algorithm. 

\begin{figure} 
    \centering
    \begin{adjustbox}{width=\columnwidth}
\begin{tikzpicture}

\definecolor{color0}{rgb}{0.12156862745098,0.466666666666667,0.705882352941177}

\begin{axis}[
tick pos=left,
xlabel={PC1},
xmin=-540.470196000102, xmax=1001.94463272543,
ylabel={PC2},
ymin=-31.6833319156906, ymax=63.4279488182277
]
\addplot [only marks, draw=color0, fill=color0, opacity=0.5, colormap/viridis]
table{%
x                      y
317.266831985751 21.8603439511454
301.806004433014 -4.9096427600398
436.768347493708 -6.23619074934878
731.946520535103 0.551898133542204
-12.8679251704883 18.7245692848386
701.937935217716 0.0607211782508798
541.680438780814 -13.1243581035175
547.10644462508 11.7861659440381
296.74537043374 -7.78357992885779
296.757635296565 -6.74818638530355
761.786938281927 -8.00320727686149
531.651762626183 -13.9392862731823
571.544083970909 -20.6409373800114
401.635457948544 -15.6504891450875
798.762427298232 -11.6551539463816
561.951718021998 2.56125099075781
532.083967087165 11.0897589404576
382.022469637855 8.75310345398118
931.825907209886 -7.9789456028511
97.1083321644641 14.757793559292
32.2971164927613 25.9139357736638
21.8527723373736 2.10249882129662
286.800343012639 -3.62813601610314
266.689421659569 -9.26210267905242
96.7269518899732 -5.24395205638919
82.2006816695031 23.0427747757108
446.634460426242 -14.4438652112957
536.627747268729 -15.0461699136749
166.916206422832 4.52744689666769
286.716434366017 -8.60804065752158
536.733053997797 -7.99051244505336
766.799867230677 -7.06907524871458
241.857423540708 0.180625741698097
487.303834830015 23.8607685120415
346.936599530058 4.33153761474323
171.78653765138 -2.54504918648976
131.992445734786 8.13156706644268
356.726125168147 -7.84520214250648
271.752365965244 -6.36402162883768
12.3466582103291 28.2550150049297
47.1344903889274 16.661999923353
286.592518625669 -14.5945890987717
346.803182127675 -4.67079801229819
-68.108377056805 4.69072860830196
136.933220257529 5.05392023046133
331.960247569762 5.61722157927964
316.819196059385 -3.13423522636249
236.818989660809 -2.70262047244776
311.826636048791 -2.00872839515736
511.894473512462 -0.524753003425865
401.648834388582 -14.618643253673
516.637599855805 -14.6628820538675
441.973214131087 3.66375715844714
626.991830599307 4.39181696785985
312.102612008708 12.960360345693
372.037863025269 9.9291008394944
222.122036227963 14.5658458733339
521.780428291889 -6.75067699701962
536.888918661199 -1.00728557775475
-68.1433024882394 2.6635629403769
-298.117832896025 5.78882007045659
-118.263041200239 -3.43228425086219
-328.351050297887 -6.69795405616434
-393.04716508925 11.4313858711039
-70.1987833204788 -0.268749984129148
-246.502168468175 -17.1531294544537
-238.532065005798 -17.2851864979501
2.00157513724531 10.4220966306491
-29.2667356191633 51.9668640691108
121.836012274972 1.31878460778751
-338.397198914897 -7.50343297879075
-276.389011019471 -7.59908025034458
137.800242338884 -0.974664696073356
-320.169261331115 3.14027260879853
-356.351528354769 -7.20698199588571
-247.929118612596 16.8508506084143
2.47377229339041 36.4085008637925
-285.137171514553 6.53311390362017
-470.351470484556 -5.21461549603527
-34.4227490652938 -12.9152718354486
-118.581902909118 -19.4248882886601
-233.425354899068 -10.3357633433576
-228.246022993864 -1.49611637536515
-298.145743690276 4.7329191706756
-253.339615692447 -5.03338344095827
-186.402324208497 -8.20892424219682
-68.4725836331658 -14.3138046285345
-123.721613858941 -27.3462949728384
-268.476418035196 -13.7837320611844
-298.396919107266 -8.21821159917473
-253.505208930766 -15.0255575441508
-458.314484559009 -3.42998721262897
-403.144844754059 5.60486218820666
189.881070910981 59.0909118753755
-122.578446211626 36.6465890267283
-320.379416907007 -8.87619281970133
-88.3718881119528 -9.95671437250267
-342.332971978403 -5.48071542756554
-38.2244347785674 -1.86663246112257
-186.365694164277 -8.24403070046581
-310.176472400156 3.96530566051935
-333.381398722882 -7.63997324733296
-76.4448701183074 -13.1813044762534
-433.303065054956 -1.84569101083536
-238.497610390075 -15.3037722175237
-260.440626672515 -10.8923454321395
-436.250742623602 0.19017089056254
-68.2835851541939 -4.33298254108343
-186.031372155472 10.7683263255112
-423.335691169594 -4.03950573183251
-141.113021184297 5.98515009223289
-314.353522338886 -5.95551359250125
-363.422000854469 -9.08573723179191
-341.407350161369 -8.49566854004658
-253.401721728647 -9.05827463365556
-402.991978628193 15.6058617848431
-376.461696201015 -12.8593828366196
-184.389958804257 -9.29277881651395
-123.238508015346 -1.34976883217545
-383.122758599743 9.2825949887517
-368.381980776913 -6.98998481193257
-368.44900858671 -10.9950488336466
-370.395836970135 -7.971125762252
-396.371257734007 -6.48616782945864
-282.325027135277 -2.48350858601439
-406.354748553224 -4.3219182447417
-168.529185411753 -16.5326068227259
-117.769698049223 24.5739334444792
-218.07913654697 8.38955197133459
-188.209562023789 1.88321098454373
-148.064652944425 9.14981655945354
-98.424077302299 -12.7460480352402
-53.2745644387561 -4.51210848777824
-28.3904229851905 -9.98742512902379
-233.246099426715 0.684151592766971
-168.368358300269 -8.48573112742976
-158.162949411782 4.33496547217101
-148.240071738843 -0.860073931225779
31.6122475878958 -11.0382580589717
-228.220577801005 1.56861383736767
-198.297272179513 -3.97273087512042
107.01776658258 10.6657658297275
81.8240747112375 1.05715210204834
-333.487218355873 -13.5595333394853
-123.418176435756 -11.2378838878097
-98.3122304369104 -5.67107254626927
-197.92081930713 17.1114475932034
-247.747255635159 27.9922220963859
-267.924747713031 17.375143654705
-322.868215850454 22.3032609622543
-73.1906409259249 -0.0924694983307521
-108.110777351487 5.47447080469659
-23.3133327454909 -5.99967729278506
-268.323972370119 -5.65079463632483
-128.35369887178 -8.07854133094033
-228.36121900552 -7.38289697010796
-68.0502589146213 8.73971213551171
-178.060350368794 9.69176321257319
-73.0628200543294 7.81075709638775
-133.340887822421 -7.02866815542043
-228.370629076877 -7.38660441559712
-52.9852687145031 12.5730196332565
-63.376741218139 -10.2705930863695
1.88759172693318 5.56973878982697
-117.971344975726 14.6971377534775
-238.214488028722 0.742225748558446
-278.375317421062 -8.46744208188486
-88.3197040219976 -6.83322684077667
-8.27328199596616 -4.27726554867408
1.83706295291546 2.54413877237885
87.1617417212274 19.0661119146614
92.1599266484904 18.9571239175922
-188.237334535129 -0.0431841055536118
};
\addplot [only marks, draw=red, fill=red, colormap/viridis]
table{%
x                      y
237.415737549942 35.302848097667
-228.323193948353 -7.54525939454033
-88.2080963949882 0.244321431980082
131.705446608074 -4.74223128597482
-282.832882774938 24.5564334521563
};
\addplot [semithick, color0]
table {%
-540.470196000102 -7.105427357601e-15
1001.94463272543 -7.105427357601e-15
};
\addplot [semithick, color0]
table {%
-2.27373675443232e-13 -31.6833319156906
-2.27373675443232e-13 63.4279488182277
};
\draw[] (axis cs:237.415737549942,35.302848097667) -- (axis cs:237.415737549942,35.302848097667);
\node at (axis cs:237.415737549942,35.302848097667)[
  scale=0.5,
  anchor=base west,
  text=black,
  rotate=0.0
]{73};
\draw[] (axis cs:-228.323193948353,-7.54525939454033) -- (axis cs:-228.323193948353,-7.54525939454033);
\node at (axis cs:-228.323193948353,-7.54525939454033)[
  scale=0.5,
  anchor=base west,
  text=black,
  rotate=0.0
]{59};
\draw[] (axis cs:-88.2080963949882,0.244321431980082) -- (axis cs:-88.2080963949882,0.244321431980082);
\node at (axis cs:-88.2080963949882,0.244321431980082)[
  scale=0.5,
  anchor=base west,
  text=black,
  rotate=0.0
]{158};
\draw[] (axis cs:131.705446608074,-4.74223128597482) -- (axis cs:131.705446608074,-4.74223128597482);
\node at (axis cs:131.705446608074,-4.74223128597482)[
  scale=0.5,
  anchor=base west,
  text=black,
  rotate=0.0
]{157};
\draw[] (axis cs:-282.832882774938,24.5564334521563) -- (axis cs:-282.832882774938,24.5564334521563);
\node at (axis cs:-282.832882774938,24.5564334521563)[
  scale=0.5,
  anchor=base west,
  text=black,
  rotate=0.0
]{121};
\end{axis}

\end{tikzpicture}
\begin{tikzpicture}

\definecolor{color0}{rgb}{0.12156862745098,0.466666666666667,0.705882352941177}

\begin{axis}[
tick pos=left,
xlabel={PC1},
xmin=-540.470196000102, xmax=1001.94463272543,
ylabel={PC3},
ymin=-10.9032984697831, ymax=10.2345779296305
]
\addplot [only marks, draw=color0, fill=color0, opacity=0.5, colormap/viridis]
table{%
x                      y
317.266831985751 -2.57578578920518
301.806004433014 -6.57500158512211
436.768347493708 1.06166950567778
731.946520535103 1.15807650514416
-12.8679251704883 0.833288182541045
701.937935217716 -0.674874989195406
541.680438780814 -2.10399409651874
547.10644462508 0.298494163066539
296.74537043374 -3.74271354719
296.757635296565 -1.42482900482689
761.786938281927 1.83532946483522
531.651762626183 -0.199432863659044
571.544083970909 -0.521489264734274
401.635457948544 -5.61440504870796
798.762427298232 -2.83176626765209
561.951718021998 0.836348594994822
532.083967087165 2.7480773502469
382.022469637855 2.21801406751185
931.825907209886 2.0969095556739
97.1083321644641 -3.59485734372519
32.2971164927613 -3.42296930202204
21.8527723373736 -0.740901130977805
286.800343012639 -2.08381347199213
266.689421659569 -0.903441968274234
96.7269518899732 0.186298502889002
82.2006816695031 4.37550751614166
446.634460426242 -1.33726116587159
536.627747268729 -0.317464477581158
166.916206422832 0.152912838011959
286.716434366017 -2.15855832155734
536.733053997797 4.9114971437779
766.799867230677 3.22665441293051
241.857423540708 -1.61430897297771
487.303834830015 1.63014341740798
346.936599530058 0.548218616720375
171.78653765138 1.42042793966059
131.992445734786 -3.38805454491149
356.726125168147 -0.134379891220054
271.752365965244 -3.01635599264162
12.3466582103291 -5.78623260691637
47.1344903889274 -2.89817210187545
286.592518625669 0.661266308301254
346.803182127675 -2.74642355340049
-68.108377056805 -2.10872587700785
136.933220257529 -2.01691957100457
331.960247569762 1.0530188207575
316.819196059385 -1.85982510401619
236.818989660809 -2.09344798733299
311.826636048791 0.954458039606144
511.894473512462 1.26149378482361
401.648834388582 -4.10226737513046
516.637599855805 0.158432199083612
441.973214131087 -2.86828070160287
626.991830599307 0.640708886417213
312.102612008708 -1.574388519465
372.037863025269 2.4962700150435
222.122036227963 -1.91828738061502
521.780428291889 -0.0703848200496018
536.888918661199 -0.0754231729250994
-68.1433024882394 -3.86219322347913
-298.117832896025 -3.21066912303333
-118.263041200239 -2.16393931771962
-328.351050297887 -2.26872070047846
-393.04716508925 -2.80938174882973
-70.1987833204788 -1.84432617871848
-246.502168468175 -5.13570805863723
-238.532065005798 -0.937412371472312
2.00157513724531 -2.893128720421
-29.2667356191633 -4.04335349000603
121.836012274972 0.86659026853359
-338.397198914897 2.86502578335421
-276.389011019471 2.65260344764041
137.800242338884 0.982250156899061
-320.169261331115 -4.72264801044095
-356.351528354769 -4.72055905797996
-247.929118612596 -3.03112080882931
2.47377229339041 -5.33987055078557
-285.137171514553 1.21849711216235
-470.351470484556 -3.41874992693811
-34.4227490652938 -1.1236446846823
-118.581902909118 2.76266673350521
-233.425354899068 2.3585559281542
-228.246022993864 -3.18129381497937
-298.145743690276 -3.60507834040079
-253.339615692447 1.17848679502251
-186.402324208497 4.31632304749533
-68.4725836331658 1.09845929977996
-123.721613858941 2.27916402320357
-268.476418035196 -2.56453871245193
-298.396919107266 0.764621759292739
-253.505208930766 -0.26903956316685
-458.314484559009 -4.12138429224773
-403.144844754059 -3.75085623101417
189.881070910981 -1.46019659876743
-122.578446211626 0.252283546878175
-320.379416907007 -5.24410185665738
-88.3718881119528 -1.79431486647003
-342.332971978403 -3.65250114207064
-38.2244347785674 -2.57229388009484
-186.365694164277 -2.38488175913226
-310.176472400156 -0.703039378704368
-333.381398722882 -2.26130444302599
-76.4448701183074 -0.082273619197417
-433.303065054956 -0.146294940624469
-238.497610390075 -1.8189610112132
-260.440626672515 1.31060722045756
-436.250742623602 -3.18044068483487
-68.2835851541939 -0.95707455399146
-186.031372155472 -1.60932396027221
-423.335691169594 -1.30702153710039
-141.113021184297 -0.384440801958473
-314.353522338886 -0.683718465897873
-363.422000854469 0.450707806015194
-341.407350161369 -0.766232894905433
-253.401721728647 -0.899546313006856
-402.991978628193 -0.219462242914414
-376.461696201015 -4.51800730065423
-184.389958804257 -2.13685431928623
-123.238508015346 -0.814641704495966
-383.122758599743 3.9783914592289
-368.381980776913 0.0504736492015942
-368.44900858671 -0.723804886320857
-370.395836970135 -0.951546431570072
-396.371257734007 -0.334924433273943
-282.325027135277 6.31166626942994
-406.354748553224 1.77829643602875
-168.529185411753 1.39776380269821
-117.769698049223 -2.13884149000489
-218.07913654697 0.20009681297066
-188.209562023789 4.01284071370572
-148.064652944425 1.73945165049282
-98.424077302299 -1.57741756435062
-53.2745644387561 0.209834797539657
-28.3904229851905 1.91315569471647
-233.246099426715 4.99276267044973
-168.368358300269 0.408639801421746
-158.162949411782 3.83608550432627
-148.240071738843 1.18039742304747
31.6122475878958 1.43621859150828
-228.220577801005 3.0909690910993
-198.297272179513 0.343749070033149
107.01776658258 1.21920230875675
81.8240747112375 2.00561747785688
-333.487218355873 0.3632379882177
-123.418176435756 3.24373994360802
-98.3122304369104 3.31068644456946
-197.92081930713 3.0673649056019
-247.747255635159 4.55889421184437
-267.924747713031 3.50724002785788
-322.868215850454 4.89663274962588
-73.1906409259249 1.38946821147499
-108.110777351487 1.19848804998739
-23.3133327454909 4.21916341270019
-268.323972370119 1.19640003079023
-128.35369887178 4.90162586628323
-228.36121900552 2.12322325165832
-68.0502589146213 1.04963094387148
-178.060350368794 2.23127728533122
-73.0628200543294 -0.498115008237464
-133.340887822421 4.0331516109786
-228.370629076877 3.32845759827156
-52.9852687145031 5.48629387867153
-63.376741218139 2.41118850069184
1.88759172693318 6.23311773022001
-117.971344975726 6.32926022313749
-238.214488028722 -0.596167796896381
-278.375317421062 1.30547401433324
-88.3197040219976 2.40574035414687
-8.27328199596616 2.94290455505108
1.83706295291546 4.67280038523307
87.1617417212274 3.2326579573789
92.1599266484904 2.64446934636677
-188.237334535129 6.02223916247466
};
\addplot [only marks, draw=red, fill=red, colormap/viridis]
table{%
x                      y
237.415737549942 9.02039272240476
-228.323193948353 -9.68911326255737
-88.2080963949882 7.59318099714902
131.705446608074 8.77978141003079
-282.832882774938 6.43480080348146
};
\addplot [semithick, color0]
table {%
-540.470196000102 0
1001.94463272543 0
};
\addplot [semithick, color0]
table {%
-2.27373675443232e-13 -10.9032984697831
-2.27373675443232e-13 10.2345779296305
};
\draw[] (axis cs:237.415737549942,9.02039272240476) -- (axis cs:237.415737549942,9.02039272240476);
\node at (axis cs:237.415737549942,9.02039272240476)[
  scale=0.5,
  anchor=base west,
  text=black,
  rotate=0.0
]{73};
\draw[] (axis cs:-228.323193948353,-9.68911326255737) -- (axis cs:-228.323193948353,-9.68911326255737);
\node at (axis cs:-228.323193948353,-9.68911326255737)[
  scale=0.5,
  anchor=base west,
  text=black,
  rotate=0.0
]{59};
\draw[] (axis cs:-88.2080963949882,7.59318099714902) -- (axis cs:-88.2080963949882,7.59318099714902);
\node at (axis cs:-88.2080963949882,7.59318099714902)[
  scale=0.5,
  anchor=base west,
  text=black,
  rotate=0.0
]{158};
\draw[] (axis cs:131.705446608074,8.77978141003079) -- (axis cs:131.705446608074,8.77978141003079);
\node at (axis cs:131.705446608074,8.77978141003079)[
  scale=0.5,
  anchor=base west,
  text=black,
  rotate=0.0
]{157};
\draw[] (axis cs:-282.832882774938,6.43480080348146) -- (axis cs:-282.832882774938,6.43480080348146);
\node at (axis cs:-282.832882774938,6.43480080348146)[
  scale=0.5,
  anchor=base west,
  text=black,
  rotate=0.0
]{121};
\end{axis}

\end{tikzpicture}
    \end{adjustbox}
    \caption{Outliers selected by our COPT algorithm
    on the wine dataset. (Red points are the outliers,
    $k {=} 5$, $r {=} 2$
    $E_\text{rpca} {=} 14.7220$).}
    \label{fig:A*outliers}
\end{figure}
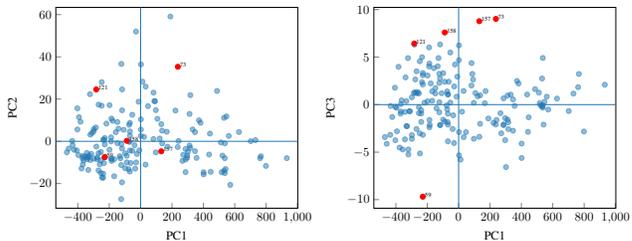

\begin{figure} 
    \centering
    \begin{adjustbox}{width=\columnwidth}
\begin{tikzpicture}

\definecolor{color0}{rgb}{0.12156862745098,0.466666666666667,0.705882352941177}

\begin{axis}[
tick pos=left,
xlabel={PC1},
xmin=-538.981747273402, xmax=1003.4356220359,
ylabel={PC2},
ymin=-31.1170870082449, ymax=65.4378443392539
]
\addplot [only marks, draw=color0, fill=color0, opacity=0.5, colormap/viridis]
table{%
x                      y
318.741756076429 22.6936705209738
303.293613628301 -4.1273351536069
438.258204355198 -5.30028471083659
733.432932015998 1.66091631185589
-11.3906772457441 19.4228843340862
703.42418845751 1.12159975004338
543.173078116123 -12.1689975718427
548.586962791414 12.7538664577868
298.23502873978 -6.95741182446327
298.24729513431 -5.87522930527462
763.277764926058 -6.90399815813195
533.145179625471 -12.9753300088611
573.040841571639 -19.6492153500813
403.128715509535 -14.7875204279502
800.254168228723 -10.5554752773328
563.43704184257 3.57746066943632
533.56536418494 12.0916677920389
383.504988888224 9.68613587289728
933.316781039275 -6.75141023633845
98.5866738045553 15.4829787661771
33.7698859041923 26.6061262551805
23.3381046057662 2.81423166918736
288.288244415894 -2.81867033159132
268.180400514475 -8.45045428939604
98.2161726575866 -4.50474611352728
83.6764457765213 23.8013732209209
448.127948789021 -13.5314235085703
538.121676673203 -14.0987429748269
168.400475953551 5.31225610514043
288.206829460639 -7.78477240699166
538.224555172423 -6.96953944159641
768.290507700751 -5.94001993794727
243.343474770392 0.971478328895188
488.778533237041 24.8095273811705
348.420966633332 5.19585393386559
173.274651394461 -1.73704991877712
133.474144838297 8.86411483349491
358.216502288978 -6.97799419302086
273.241420742817 -5.57366056439644
13.817748144345 28.9183540282346
48.6120521865688 17.3762724008513
288.086509246357 -13.7376139567214
348.29149775655 -3.80996379999446
-66.6246231548492 5.34904526332873
138.416788031394 5.80538027568906
333.444100695704 6.50920179541178
318.306904578973 -2.28162749368776
238.30646828968 -1.88154010287806
313.314346605365 -1.12202797740656
513.381503825635 0.497006344585736
403.141891021327 -13.7256737770041
518.131505342482 -13.6950398554966
443.45728726423 4.58661623405266
628.476158089883 5.42007228444935
313.582222963706 13.8090189244817
373.519838752974 10.8481958356995
223.600808003967 15.3786284442753
523.270282500035 -5.77677350025072
538.375888965414 -0.0182294481575025
-226.834901002364 -7.08196016029353
-66.6589834976046 3.27465289682902
-296.634864293568 6.33832262377862
-116.775230343166 -2.8115168054629
-326.861477383743 -6.16976129239399
-391.566953235576 11.8927366597818
-68.7124738323757 0.379882738770412
-245.007900170822 -16.5968309834707
-237.036916710492 -16.6958839528485
3.48216397033794 11.0742967081848
123.321930570781 2.05557684658442
-336.90617527848 -6.93869934225129
-274.898049992029 -7.00381876177356
139.287428799574 -0.226370348007667
-318.685267646665 3.63121066437536
-354.862252854126 -6.71376229696571
-246.451746863716 17.3728769769063
3.94073559921748 37.0115222250114
-283.653627862592 7.09044745441077
-468.862906276775 -4.79804628001273
-32.929899329015 -12.2440917802103
-117.084996969512 -18.7837272620459
-231.933088839888 -9.70031724863704
-226.759342489568 -0.96249613683665
-296.662304002202 5.22651254903097
-251.85029240842 -4.47126456646022
-184.910741528629 -7.58389444286288
-66.9786099123399 -13.656231657928
-122.220791031812 -26.7144295829572
-266.98343900343 -13.2601852424914
-296.90604798987 -7.66444537388559
-252.011133128672 -14.4614212401084
-456.826966981748 -3.01256412942779
-401.661828942177 6.06017742863769
-121.110461980417 37.2390540025156
-318.889407002422 -8.39647965819341
-86.8806133518276 -9.30999335184039
-340.844328765783 -5.00168035996552
-36.7375019147086 -1.22819859644429
-184.875501500609 -7.68396308845719
-308.69202801959 4.49143402178318
-331.891442721776 -7.15235866549675
-74.9516838743896 -12.5365248590617
-431.81555333094 -1.37126803548116
-237.003669195903 -14.7432625458403
-258.948296344722 -10.3133423631066
-434.764877285426 0.624710648579199
-66.7950263920087 -3.70659854483071
-184.550572900309 11.3335215920471
-421.847312469212 -3.57766746095151
-139.629592040685 6.60614201211586
-312.864074119295 -5.43355924550405
-361.93069866271 -8.56892403907204
-339.916622898227 -8.00618787520189
-251.910774398554 -8.52506135423464
-401.513328410204 16.0704412041945
-374.969563674203 -12.3762116207419
-182.899176677498 -8.74083955667928
-121.751413095853 -0.733814355609415
-381.640037607821 9.81379639879962
-366.891812623602 -6.46443687957913
-366.956937512098 -10.4866446519434
-368.905375677464 -7.47145786552099
-394.881390628012 -5.97603875828922
-280.835896750884 -1.87612868511132
-404.8655883515 -3.82798570757537
-167.034005483417 -15.907507256711
-116.296097845962 25.181591445355
-216.596820214594 9.00918780744519
-186.723178246875 2.55694833244405
-146.58242753751 9.8118485651044
-96.931465791985 -12.0888189703059
-51.7857451332305 -3.78983512110585
-26.8985437450251 -9.27419882145493
-231.758900648426 1.34810334819426
-166.877475378642 -7.82490267226239
-156.677843147364 5.01054752638695
-146.752895913429 -0.208156917601781
33.1045938798142 -10.2805314028152
-226.734226859114 2.20229202219212
-196.808676260308 -3.340944935082
108.499096036731 11.4834569487696
83.3103688663748 1.81422641027611
-331.993767111978 -12.9754176456264
-121.925294897987 -10.4952219041846
-96.8221477391119 -4.91272012357212
-196.442275250457 17.8278331023735
-246.273903242801 28.6810517360076
-266.446205792282 18.0773982291954
-321.391911303978 22.9330028859804
-71.703754987778 0.685061548874899
-106.626791397558 6.17533525199161
-21.8229300466439 -5.19977591484933
-266.83428051831 -4.9767184540582
133.196111736534 -3.84533588234951
-126.861990988176 -7.27594632238525
-226.870484851873 -6.70640040765074
-66.5679800026794 9.44735700914869
-176.578278976269 10.3500698228468
-71.5803989510484 8.48934811028962
-131.849953540935 -6.26176837439706
-226.879677408759 -6.70913413827743
-51.503951046519 13.3994622781522
-61.8845066648589 -9.47636422814028
3.37256993136232 6.39658257372385
-116.490929243419 15.4738463474376
-236.728470851769 1.34561222790948
-276.884182730243 -7.79237085854802
-86.8292064544031 -6.05851826020486
-6.78401738330458 -3.47984766155896
3.32322610691298 3.34279083525674
88.639255988384 19.9259028700544
93.637364971855 19.791810394123
-186.749505218211 0.716637586082583
};
\addplot [only marks, draw=red, fill=red, colormap/viridis]
table{%
x                      y
-281.357253245756 25.201807375898
191.337322581713 59.7954085291894
-27.8073815941288 52.5492036637691
238.886224074424 36.1636547268308
-86.7200266292095 1.10836352800612
};
\addplot [semithick, color0]
table {%
-538.981747273402 0
1003.4356220359 0
};
\addplot [semithick, color0]
table {%
-1.13686837721616e-13 -31.1170870082449
-1.13686837721616e-13 65.4378443392539
};
\draw[] (axis cs:-281.357253245756,25.201807375898) -- (axis cs:-281.357253245756,25.201807375898);
\node at (axis cs:-281.357253245756,25.201807375898)[
  scale=0.5,
  anchor=base west,
  text=black,
  rotate=0.0
]{121};
\draw[] (axis cs:191.337322581713,59.7954085291894) -- (axis cs:191.337322581713,59.7954085291894);
\node at (axis cs:191.337322581713,59.7954085291894)[
  scale=0.5,
  anchor=base west,
  text=black,
  rotate=0.0
]{95};
\draw[] (axis cs:-27.8073815941288,52.5492036637691) -- (axis cs:-27.8073815941288,52.5492036637691);
\node at (axis cs:-27.8073815941288,52.5492036637691)[
  scale=0.5,
  anchor=base west,
  text=black,
  rotate=0.0
]{69};
\draw[] (axis cs:238.886224074424,36.1636547268308) -- (axis cs:238.886224074424,36.1636547268308);
\node at (axis cs:238.886224074424,36.1636547268308)[
  scale=0.5,
  anchor=base west,
  text=black,
  rotate=0.0
]{73};
\draw[] (axis cs:-86.7200266292095,1.10836352800612) -- (axis cs:-86.7200266292095,1.10836352800612);
\node at (axis cs:-86.7200266292095,1.10836352800612)[
  scale=0.5,
  anchor=base west,
  text=black,
  rotate=0.0
]{158};
\end{axis}

\end{tikzpicture}
\begin{tikzpicture}

\definecolor{color0}{rgb}{0.12156862745098,0.466666666666667,0.705882352941177}

\begin{axis}[
tick pos=left,
xlabel={PC1},
xmin=-538.981747273402, xmax=1003.4356220359,
ylabel={PC3},
ymin=-10.7030731649891, ymax=10.273261690786
]
\addplot [only marks, draw=color0, fill=color0, opacity=0.5, colormap/viridis]
table{%
x                      y
318.741756076429 -3.00056746677094
303.293613628301 -6.70392259962582
438.258204355198 1.11487063209137
733.432932015998 0.94751211429557
-11.3906772457441 0.643497514207015
703.42418845751 -0.881839522980263
543.173078116123 -2.07686908362259
548.586962791414 0.0692642324313594
298.23502873978 -3.81595320122622
298.24729513431 -1.53074656867193
763.277764926058 1.88187835686362
533.145179625471 -0.0989791593547778
573.040841571639 -0.367640493750877
403.128715509535 -5.63703505455068
800.254168228723 -2.98891219641547
563.43704184257 0.610641488593496
533.56536418494 2.53834080606088
383.504988888224 2.02876183277549
933.316781039275 1.94559153114549
98.5866738045553 -3.93981321759114
33.7698859041923 -3.89746329498808
23.3381046057662 -0.784736064486077
288.288244415894 -2.03333187681809
268.180400514475 -0.751367188600717
98.2161726575866 0.363514103434505
83.6764457765213 4.29947109271923
448.127948789021 -1.21477680725514
538.121676673203 -0.136429745056359
168.400475953551 0.121252100065898
288.206829460639 -2.12121446184653
538.224555172423 5.0906997364182
768.290507700751 3.2281929995164
243.343474770392 -1.62307880824362
488.778533237041 1.2793518620806
348.420966633332 0.508319318044762
173.274651394461 1.49294058947141
133.474144838297 -3.60059600500219
358.216502288978 -0.0374463216227095
273.241420742817 -2.97500222933363
13.817748144345 -6.38626460247648
48.6120521865688 -3.25870749761618
288.086509246357 0.84820727085666
348.29149775655 -2.81249160525871
-66.6246231548492 -2.23206271010011
138.416788031394 -2.137415502241
333.444100695704 0.901239107248685
318.306904578973 -1.92252788950468
238.30646828968 -2.19548896545973
313.314346605365 0.893520163105648
513.381503825635 1.01582133212618
403.141891021327 -4.18733884314619
518.131505342482 0.283060636909314
443.45728726423 -3.15754824597398
628.476158089883 0.442037641749037
313.582222963706 -1.88591955936428
373.519838752974 2.33544135947757
223.600808003967 -2.27904777077061
523.270282500035 -0.0973601566508276
538.375888965414 -0.211551761774746
-226.834901002364 -9.73580648932522
-66.6589834976046 -3.95927222319852
-296.634864293568 -3.47094561714438
-116.775230343166 -2.14052653116699
-326.861477383743 -2.13058345226161
-391.566953235576 -2.90026664694181
-68.7124738323757 -1.84874078685478
-245.007900170822 -5.0133024975257
-237.036916710492 -0.677294139508491
3.48216397033794 -3.06488495289522
123.321930570781 0.932739066037183
-336.90617527848 3.21383993918022
-274.898049992029 2.91212437672466
139.287428799574 1.16517403003997
-318.685267646665 -4.83419533893291
-354.862252854126 -4.72037894026124
-246.451746863716 -3.22793244054845
3.94073559921748 -5.88128711477459
-283.653627862592 1.34063637366863
-468.862906276775 -3.21057568343378
-32.929899329015 -0.900581099320759
-117.084996969512 3.30529989651858
-231.933088839888 2.51915142181685
-226.759342489568 -3.06762418924188
-296.662304002202 -3.58446312581481
-251.85029240842 1.45790470969548
-184.910741528629 4.7464174325186
-66.9786099123399 1.46357309770008
-122.220791031812 2.95175431177948
-266.98343900343 -2.29090911390603
-296.90604798987 1.0029737703501
-252.011133128672 0.0365822056937175
-456.826966981748 -3.96743698458478
-401.661828942177 -3.7710004361548
-121.110461980417 -0.097976239991938
-318.889407002422 -5.10008535183549
-86.8806133518276 -1.61368383285771
-340.844328765783 -3.46108612649264
-36.7375019147086 -2.51445497404259
-184.875501500609 -2.18885638976183
-308.69202801959 -0.606175177366002
-331.891442721776 -2.01285938016962
-74.9516838743896 0.242541267535785
-431.81555333094 0.0499365726322311
-237.003669195903 -1.53921150460955
-258.948296344722 1.60520560217732
-434.764877285426 -3.06440235402387
-66.7950263920087 -0.702819961828505
-184.550572900309 -1.6205615968723
-421.847312469212 -1.07042551947341
-139.629592040685 -0.416211528356345
-312.864074119295 -0.446086684409818
-361.93069866271 0.779162144508742
-339.916622898227 -0.394384936624081
-251.910774398554 -0.5683720012221
-401.513328410204 -0.197341598660685
-374.969563674203 -4.44053620263109
-182.899176677498 -1.84606860657546
-121.751413095853 -0.657337662187868
-381.640037607821 4.16786286084041
-366.891812623602 0.278828478739813
-366.956937512098 -0.392576432260106
-368.905375677464 -0.671932603731045
-394.881390628012 -0.116940883266074
-280.835896750884 6.70373656257118
-404.8655883515 2.13779755145581
-167.034005483417 1.77366942280512
-116.296097845962 -2.54211418835204
-216.596820214594 -0.0201975479200574
-186.723178246875 3.986856803704
-146.58242753751 1.56988386727681
-96.931465791985 -1.54512607999501
-51.7857451332305 0.0438409731476766
-26.8985437450251 2.03359342255003
-231.758900648426 5.0164830372718
-166.877475378642 0.371817097646394
-156.677843147364 3.82598232117189
-146.752895913429 1.16913069168453
33.1045938798142 1.48179107904309
-226.734226859114 3.10846056607046
-196.808676260308 0.323647782545573
108.499096036731 0.782399473461197
83.3103688663748 1.98621810590256
-331.993767111978 0.404440704889196
-121.925294897987 3.19413897957727
-96.8221477391119 3.15944661390585
-196.442275250457 2.54042772802027
-246.273903242801 4.00482744461597
-266.446205792282 2.93919350869007
-321.391911303978 4.57465019138564
-71.703754987778 0.969088011174657
-106.626791397558 0.929603678680234
-21.8229300466439 4.09010298939563
-266.83428051831 0.937249374091078
133.196111736534 8.84222209075468
-126.861990988176 4.67865027230957
-226.870484851873 2.02960627437426
-66.5679800026794 0.78487625571666
-176.578278976269 2.04575966915705
-71.5803989510484 -0.752334872234766
-131.849953540935 3.84323974062144
-226.879677408759 3.29631315246161
-51.503951046519 4.99901439086065
-61.8845066648589 2.16621991485851
3.37256993136232 6.0139906551068
-116.490929243419 5.97979618572447
-236.728470851769 -0.741601387330336
-276.884182730243 1.05495493833456
-86.8292064544031 2.15364823938477
-6.78401738330458 2.74008326943738
3.32322610691298 4.50443107075273
88.639255988384 2.58563895732221
93.637364971855 2.07710159520684
-186.749505218211 5.81205587206593
};
\addplot [only marks, draw=red, fill=red, colormap/viridis]
table{%
x                      y
-281.357253245756 6.37540912415312
191.337322581713 -2.13103258009744
-27.8073815941288 -4.71176449580637
238.886224074424 8.95525369674856
-86.7200266292095 7.25997129445666
};
\addplot [semithick, color0]
table {%
-538.981747273402 1.77635683940025e-15
1003.4356220359 1.77635683940025e-15
};
\addplot [semithick, color0]
table {%
-1.13686837721616e-13 -10.7030731649891
-1.13686837721616e-13 10.273261690786
};
\draw[] (axis cs:-281.357253245756,6.37540912415312) -- (axis cs:-281.357253245756,6.37540912415312);
\node at (axis cs:-281.357253245756,6.37540912415312)[
  scale=0.5,
  anchor=base west,
  text=black,
  rotate=0.0
]{121};
\draw[] (axis cs:191.337322581713,-2.13103258009744) -- (axis cs:191.337322581713,-2.13103258009744);
\node at (axis cs:191.337322581713,-2.13103258009744)[
  scale=0.5,
  anchor=base west,
  text=black,
  rotate=0.0
]{95};
\draw[] (axis cs:-27.8073815941288,-4.71176449580637) -- (axis cs:-27.8073815941288,-4.71176449580637);
\node at (axis cs:-27.8073815941288,-4.71176449580637)[
  scale=0.5,
  anchor=base west,
  text=black,
  rotate=0.0
]{69};
\draw[] (axis cs:238.886224074424,8.95525369674856) -- (axis cs:238.886224074424,8.95525369674856);
\node at (axis cs:238.886224074424,8.95525369674856)[
  scale=0.5,
  anchor=base west,
  text=black,
  rotate=0.0
]{73};
\draw[] (axis cs:-86.7200266292095,7.25997129445666) -- (axis cs:-86.7200266292095,7.25997129445666);
\node at (axis cs:-86.7200266292095,7.25997129445666)[
  scale=0.5,
  anchor=base west,
  text=black,
  rotate=0.0
]{158};
\end{axis}

\end{tikzpicture}
    \end{adjustbox}
    \caption{Outliers selected by the outlier-pursuit algorithm on the centered 
    wine dataset. (Red points are the outliers,
    $k {=} 5$, $r {=} 2$, $E_\text{rpca} {=} 15.5$).}
    \label{fig:outlier_pursuit}
\end{figure}
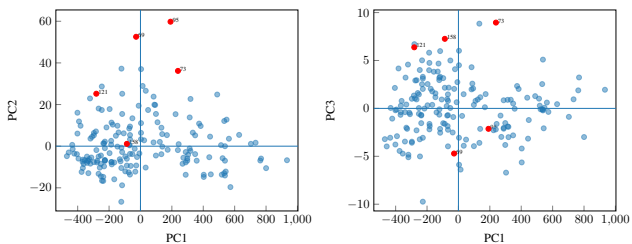

\newpage
\section*{Appendix: correctness of the bias trick}\label{appendix}
In this appendix we prove the correctness of the bias trick.
An important part can be traced back to~\cite{Cadima09}.
In that paper they prove the following result 
(as a corollary to their Theorem~2).
\begin{paragraph}{Theorem: (Cadima and Jolliffe):}
Let $B$ be the matrix of second moments of the uncentered data,
let $\mu$ be the data mean,
and let $C$ be the covariance matrix.
If one of the eigenvectors of $B$ is $\mu / \|\mu\|$ 
then all other eigenvector/eigenvalue pairs of $B$ are also 
eigenvector/eigenvalue pairs of $C$.
\end{paragraph}

\medskip

\subsection*{Notation}
Let $X = (x_1 \ldots x_n)$ be the data matrix,
let $\mu = \frac{1}{n} \sum_i x_i$ be the data mean,
and let $B = \frac{1}{n} \sum_i x_i x_i^T$ be the data
second moments matrix.
The covariance matrix is given by:
$C = \frac{1}{n} \sum_i (x_i{-}\mu) (x_i{-}\mu)^T$.
Let $\lambda_i, u_i$ be the eigenvalue/eigenvector pairs of $C$.

\noindent
Create $X_b = (x^b_1\ldots,x^b_n)$ by adding a large bias $b$ for each vector:
$x^b_i = \begin{pmatrix} x_i \\ b \end{pmatrix}$.
$X_b$ is $(m{+}1) {\times} n$.
The column mean of $X_b$ is: $\mu_b = \begin{pmatrix} \mu \\ b \end{pmatrix}$.
The corresponding $(m{+}1) {\times} (m{+}1)$ matrix of second moments is:
\[
B_b = \frac{1}{n} \sum_i^n x^b_i (x^b_i)^T =
\begin{pmatrix}
B & b \mu
\\
b \mu^T & b^2
\end{pmatrix}
\]
and the corresponding covariance matrix is:
\begin{equation} \label{Cb}
C_b = \frac{1}{n} \sum_i^n (x^b_i - \mu_b) (x^b_i - \mu_b)^T 
=
\begin{pmatrix}
C & 0
\\
0 & 0
\end{pmatrix}
\end{equation}

\noindent
Let $\lambda_i^b, u^b_i$ be the eigenvalue/eigenvector pairs of $B_b$.
Define: $\begin{pmatrix} v_i \\ w_i \end{pmatrix} = u^b_i$,
where $v_i$ is an $m$-vector and $w_i$ is a scalar.
The bias trick is useful since (as proved here) 
$u_i \approx v_{i+1}$ and $\lambda_i \approx \lambda^b_{i+1}$.
Thus, the centered eigenvectors and eigenvalues are obtained
from the uncenterd and ``biased'' eigenvectors and eigenvalues.

To analyze the bias trick we need the notion of
``approximation for sufficiently large values of the bias $b$''.
It is defined as follows:

\smallskip
\noindent
\textbf{Definition:~}
We write $p \approx q$ if for any $\epsilon > 0$ there is $b_\epsilon$
such that $(p-q)^2 < \epsilon$ for all $b > b_\epsilon$.
When $p,q$ are vectors the squared error is replaced with squared norm,
etc.
We also say ``$p$ approximates $q$'' if $p \approx q$.

\begin{paragraph}{Lemma~1.}
For a sufficiently large value of $b$:
\begin{align*}
\text{Part 1.}~ w_1 \approx \frac{b}{\sqrt{b^2 + \|\mu\|^2}}
\quad \text{Part 2.}~ v_1 \approx \frac{\mu}{\sqrt{b^2 + \|\mu\|^2}}
\end{align*}

\noindent \textbf{Proof:~}
From the Courant Fischer theorem~\cite{GV4}
the vector $v_1$ and the scalar $w_1$ minimize the following error:
\begin{multline} \label{err}
E(v_1, w_1) = \min_{a_i} \sum_i \| \begin{pmatrix}  x_i \\ b \end{pmatrix} 
- a_i \begin{pmatrix} v_1 \\ w_1 \end{pmatrix} ~\|^2
\\
= \min_{a_i} \sum_i
\|x_i - a_i v_1\|^2 + (b - a_i w_1)^2
\end{multline}
For sufficiently large value of $b$ the rightmost term dominates the error
and it is minimized by $a_i = \frac{b}{w_1}$.
Substituting this in~\refe{err} gives:
$
E(v_1,w_1) = \sum_i \|x_i - \frac{b}{w_1} v_1\|^2
$.
Since $v_1$ and $w_1$ form an eigenvector they must satisfy:
$|v_1|^2 + w_1^2 = 1$.
To minimize $E(v_1,w_1)$ subject to this constraint we use the method of
Lagrange multipliers. The Lagrangian is:
\begin{equation} \label{L}
L(v_1,w_1, \alpha) = \sum_i \|x_i - \frac{b}{w_1} v_1\|^2
+ \alpha (|v_1|^2 + w_1^2 - 1)
\end{equation}
Taking derivatives of~\refe{L} with respect to $v_1$ and equating to 0
gives:
$(-b/w_1) (n \mu - \frac{nb}{w_1} v_1) + 2 \alpha v_1 = 0$.
Therefore, the vectors $v_1$ and $\mu$ are linearly dependent
$v_1 = t \mu$. Substituting this in the constraint and solving for $t$
we get: $t = \frac{\sqrt{1-w_1^2}}{|\mu|}$,
so that:
\begin{equation} \label{v11}
v_1 \approx \frac{\sqrt{1-w_1^2}}{|\mu|} \mu
\end{equation}

\noindent
To prove Part~1 we take derivatives of~\refe{L} with respect to $w_1$
and equate to 0. This gives:
\[
 2 b n v_1^T \mu / w_1^2 - 2 n b^2 |v_1|^2 / w_1^3 + 2 \alpha w_1 =  0
\]
For sufficiently large $b$ the right most term can be ignored.
After multiplying by $w_1^3$ and simplifying this gives:
\[
 w_1 v_1^T \mu \approx b |v_1|^2 
\]
Substituting the value of $v_1$ from \refe{v11} we get the following equation
in $w_1$:
\[
w_1 \|\mu\| \approx b \sqrt{1 - w_1^2}
\]
Solving this equation for $w_1$ gives the formula in Part~1.
Substituting the Part~1 expression for $w_1$ in \refe{v11}
and simplifying gives the formula in Part~2.
\hfill $\blacksquare$
\end{paragraph}

\begin{paragraph}{Theorem~1.}
For a sufficiently large value of $b$
let $\lambda_i^b,u^b_i$ be an eigenvalue/eigenvector pair of $B_b$,
with $i{>}1$.
Suppose $u^b_i$ is partitioned as follows:
$u^b_i = \begin{pmatrix} v_i \\ w_i \end{pmatrix}$.
Then
$w_i \approx 0$ and $\lambda_i, v_i$ are approximately
eigenvalue/eigenvector pairs of $C$.

\noindent \textbf{Proof:~}
From Lemma~1 it follows that
\[
\begin{pmatrix} v_1 \\ b \end{pmatrix}
\approx \frac{ \mu_b } {\| \mu_b \|}
\]
Since this approximately satisfies the condition of
the Cadima and Jolliffe theorem 
stated above it follows that all other eigenvector/eigenvalue pairs of 
$B_b$ are also approximately eigenvector/eigenvalue pairs of $C_b$.
From \refe{Cb} it follows that 
if $z$ is the $m{+}1$ vector $\begin{pmatrix} z_1 \\ z_2 \end{pmatrix}$
where $z_1$ is an $m$-vector and $z_2$ is a scalar then
$C_b z = \begin{pmatrix} C z_1 \\ 0 \end{pmatrix}$.
Therefore, if $C_b z = \lambda z$ 
then $z_2 = 0$ and $C z_1 = \lambda z_1$.
\end{paragraph}

\begin{paragraph}{Corollary.}
The sufficiently large value of $b$ in Theorem~1
can be selected as:
\[
b \geq \frac{\sqrt{1-\epsilon^2}}{\epsilon} \| \mu \|
\]
for sufficiently small value of $\epsilon$,
where $0 \leq \epsilon \leq 1$.

\noindent \textbf{Proof:~}
Set $\epsilon = \sqrt{1 - w_1^2}$.
Substituting this value in Part~1 of Lemma~1
and solving for $b$ gives the above relation.
\hfill $\blacksquare$
\end{paragraph}

\small{

}

\end{document}